\documentclass[sigconf,nonacm,natbib=false]{acmart}

\AtBeginDocument{%
  \providecommand\BibTeX{{%
    \normalfont B\kern-0.5em{\scshape i\kern-0.25em b}\kern-0.8em\TeX}}}

\setcopyright{none}

\usepackage[
backend=biber,
style=numeric
]{biblatex}

\usepackage{float}

\graphicspath{ {images/} }
\begin{document}

\title{Community-Specific Slang and Entity Detection via Semantic Shift in Fine-Tuned Language Models}


\author{Julia Kruk}
\affiliation{%
  \institution{Georgia Institute of Technology}
  \department{College of Computing}
  \city{Atlanta}
  \state{GA}
  \country{USA}
}
\email{jkruk3@gatech.edu}

\author{Sanchita Porwal}
\affiliation{%
  \institution{Georgia Institute of Technology}
  \department{College of Computing}
  \city{Atlanta}
  \state{GA}
  \country{USA}
}
\email{sporwal9@gatech.edu}

\author{Amitrajit Bhattacharjee}
\affiliation{%
  \institution{Georgia Institute of Technology}
  \department{College of Computing}
  \city{Atlanta}
  \state{GA}
  \country{USA}
}
\email{amit.bh@gatech.edu}

\author{Mansi Phute}
\affiliation{%
  \institution{Georgia Institute of Technology}
  \department{College of Computing}
  \city{Atlanta}
  \state{GA}
  \country{USA}
}
\email{mansiphute@gatech.edu}


\begin{abstract}
We propose an unsupervised method of resolving slang, unique entities, and folklore from online communities by isolating words in the lexicon that have the highest magnitude of semantic shift. Semantic shift is defined as the evolution of a word's encoded representation as a result of fine-tuning a pretrained Large Language Model (LLM) on a community-specific text corpus. This value is inversely proportional to the cosine similarity between the base model's encoded representation of a word, and a fine-tuned model's encoded representation. We fine-tune the DistilRoBERTa model on text corpora collected from 3 Reddit subreddits (r/Technology, r/Gaming, r/WorldofWarcraft), model a distribution of cosine similarity over the lexicon, and show that one can successfully resolve words that have unique significance to the community by pulling data in the bottom 10-percentile. In contrast, we show that data in the top 10-percentile consist of words that carry relatively universal semantics.

\end{abstract}




\maketitle

\section{Introduction}




In this study, we aimed to resolve slang, unique entities, and folklore that is shared by a Reddit community in an unsupervised manner. By fine-tuning a pretrained language model on a community-specific text corpus and comparing the word embeddings of the fine-tuned model with those of the base pretrained model, we estimate magnitude of \textbf{Semantic Shift} on specific words. Semantic shift is defined as the evolution of a word's encoded representation that is a result of a community assigning unique meaning that is different from the general usage. Our hypothesis is that slang , unique entities, and folklore specific to the community will have a greater degree of change in their embedded representation as a result of the fine-tuning process, which will be reflected by lower cosine similarity values between the embeddings of the base model and the fine-tuned model. The cosine similarity is inversely proportional to the magnitude of the semantic shift.

How do we define slang, unique entities, and folklore? All of these elements are part of a distinct community's lexicon that may not be easily interpreted by the general populace or external parties. Unique entities may be individuals, organizations, or locations that have unique value to a community, such as the ASME, which is the American Society of Mechanical Engineers that was frequently referenced in the r/Technology subreddit. Folklore is focused on stories and customs that a community shares. For example, the r/WorldofWarcraft community often discusses n'Zoth, which is an old God that exists in the game. 

This community-specific approach to slang, entity, and lore detection can be used to refine state-of-the-art hate speech detection. A well-known flaw in current hate speech detection systems is that they are biased against certain communities because the slang used in these communities gets incorrectly flagged as hate speech at high frequencies (Sap et al. 2019 \cite{sap-etal-2019-risk}, Harris et al. 2021 \cite{camille-2021}). A potential application of this work can aid in de-biasing these systems, by assigning a confidence value to hate speech detection inference that is inversely proportional to a word's magnitude of semantic shift.

The use of similar language is a significant indicator of similar values and experiences. Another application of this method can be the identification of communities that have overlap in their user-base. Recent work on influence prediction (Qiu et. al. 2018 \cite{deepinf-2018}) and information flow (Bagrow et. al. 2019 \cite{bagrow-2019}) on social media platforms have strived to manually model user-base overlap by tracking specific user accounts. Yet the flow of folklore, stories, and linguistic dialects are an indicator of community interaction, and therefore can be used to hierarchically model the relationship between online communities and detect which communities have over-lapping user-bases.

\section{Related Work}

Previous works have explored various aspects of semantic change and slang detection, showcasing the potential and challenges of using these techniques for linguistic research and natural language processing. We aim to further build upon these methods to create a generalized framework to detect slang in specific communities. 

Liu et al. 2021 \cite{liu2021statistically} focuses on detecting lexical semantic change in smaller datasets, which is a challenge in historical linguistics and digital humanities. The authors propose an approach that combines contextual word embeddings with permutation-based statistical tests and uses the false discovery rate procedure to address multiple hypothesis tests. Their approach is shown to achieve high precision in simulations and real-world data from SemEval-2020 Task 1 and the Liverpool FC subreddit corpus. This work leverages semantic shift to evaluate the semantic evolution of a lexicon in the same domain over time. Where as we strive to apply this approach to understand a particular domain's deviation from the aggregate distribution. 

Marco Del Tredici et al. 2018 \cite{del2018semantic} introduce a framework for quantifying semantic variation of common words in Communities of Practice and topic-related communities. Their findings suggest that some meaning shifts are shared across related communities, while others are community-specific, providing evidence in favor of sociolinguistic theories of socially-driven semantic variation. The results are evaluated using an independent language modeling task, and the authors investigate extralinguistic features that relate to semantic variation. The authors use this approach to identify geographically located language, and leverage Word2Vec to produce vector-spaced representations of tokens.

Zhengqi Pe et al  \cite{pei2019slang} explores slang detection and identification using deep learning methods. The authors employ a combination of bidirectional recurrent neural networks, conditional random fields, and multilayer perceptrons to detect and identify slang from natural sentences. Their models perform well in sentence-level detection with an F1-score of 0.80 and token-level identification with an F1-score of 0.50. They also find that syntactic shift is a prominent feature of slang. However, an LSTM architecture which must be trained on different domains is susceptible to model shift, where the differences in word embeddings could be attributed to the randomization at initialization and training. By leveraging pretrained architectures like DistilRoBERTa, our approach is robust against this issue.

Wevers et al \cite{wevers2020digital} discusses the use of word embedding models (WEM) for tracing semantic change in historical research. They highlight the potential of WEMs to capture semantic change and assist historians in studying conceptual change or specific discursive formations over time. However, they also mention the challenges in using WEMs due to the requirement of large amounts of training data and the difficulty in evaluation. The authors examine the prerequisites for producing reliable WEMs from historical data and describe the types of research questions that can be answered using WEMs.

\begin{figure*}[ht] \includegraphics[width=\textwidth,height=6cm]{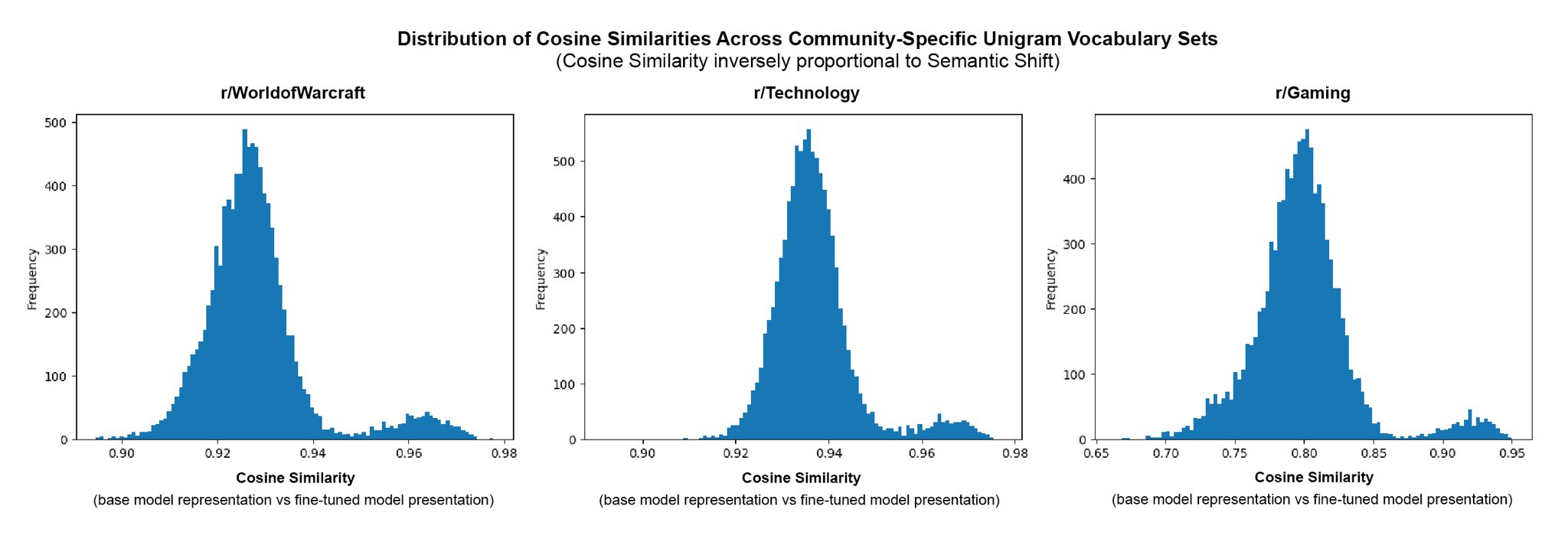}
  \caption{Semantic Shift across Reddit communities.}
\end{figure*}

\section{Dataset description}

\subsection{Data preparation}
We have separate data preparation methods for training and evaluation:
\subsubsection{Training Data} :
The dataset we used is the Reddit dataset available in the official HuggingFace library (\url{https://huggingface.co/datasets/reddit}). 
While selecting communities to analyze from the dataset, we sorted all the communities by the number of posts they contained, and we selected the communities based on their rank and relevance. We selected three communities for analysis: "World of Warcraft" (WoW), "Gaming" and "Technology" for further analysis. These datasets were chosen as they contain slang and language specific to these communities and are likely to be related to each other, allowing comparison between their results.
We extracted the content of 43,851 posts from the "Gaming" subreddit, 13,700 posts from the "Technology" subreddit, and  13,309 posts from the "World of Warcraft" subreddit. 
\subsubsection{Evaluation Data}: 
For evaluation of our task, we decided to use a "Masked Model Objective". To achieve this, we conducted a search independent from our training data, where we went online to select words from online community slang lists. We assembled a dataset consisting of 20 slang words and unique entities specific to each community and 30 slang words prevalent in the general vernacular. For each word, we provided example sentences illustrating their usage within the respective contexts. Subsequently, we masked the target words in these sentences, preparing the dataset for evaluation purposes. This approach allows for a comprehensive examination of the model's performance in identifying distinct slang, unique entities, and lore within various communities and general language use. The full evaluation set contains a total of 90 words.
\subsection{Raw Data Statistics}
The dataset has a total disk size of 22.08 GB. It contains 3,848,330 posts with an average content length of 270 words. The dataset includes fields like "author", "body", "normalizedBody", "content", "summary", "subreddit", and "subreddit id". The fields "body", "normalizedBody", and "content" all include the content of a reddit post. On analysis, "content" was the cleanest, and therefore that is what we use in all our further processing. 
\section{EXPERIMENTAL SETTINGS}
We evaluate the performance of our model on a manually created evaluation dataset. To create this dataset, we manually collected slang words from each community by browsing web pages. We also collected an example of the way the word is used in context. To prepare the data, we masked the slang word out and collected the resulting sentences as our evaluation set. Some sentences had more than one instance of the word, and in such cases, all occurrences of the word were masked. 
Evaluating such a task is not straightforward as we are implementing a new approach. However, we have still performed the following evaluation methods: perplexity, F1 score, qualitative analysis like observing filled-in masks, cosine similarity, Euclidean distance, and t-SNE visualizations.
For the training task, we primarily use Google Colab.
\section{METHODS}
\subsection{Baseline Description}
In the absence of an existing baseline for this specific task, we utilized the default DistilRoBERTa as our reference point for comparison. We selected DistilRoBERTa as the baseline model since it has been trained on masked language prediction tasks, which closely align with the objectives of our study. Additionally, we chose the distilled version of RoBERTa over RoBERTa as it is computationally much faster and has high performance. Masked language prediction tasks require the language model to comprehend the relevance of a word in relation to its surrounding words and thoroughly understand the contextual information. This principle underpins our slang detection methodology, making DistilRoBERTa an appropriate baseline for our analysis.

\subsection{Proposed method}

Through our proposed method, we aim to provide a systematic and data-driven method for identifying slang words specific to a community by leveraging the power of pre-trained language models and fine-tuning techniques.

We start with a pre-trained DistilRoBERTa model \cite{liu2019roberta} as our base model, which has been trained on a large corpus of text and is capable of generating high-quality word embeddings.

We fine-tune the base model on a dataset specific to the community of interest. This dataset should contain both slang and non-slang words that are relevant to the community. By fine-tuning the model on this dataset, we expect it to better capture the nuances and slang specific to the community.

We extract the word embeddings for a set of target words from both the base model and the fine-tuned model. These target words include slang words and non-slang words to serve as a reference for comparison.

We calculate the cosine similarity between the embeddings of each target word from the base model and the fine-tuned model. Lower cosine similarity values indicate a greater degree of change in the embeddings, which we hypothesize to be indicative of slang words specific to the community. In contrast, Words with high cosine similarity are likely to be non-slang words, as they share a similar semantic space in the embedding model.

Finally, we visualize the embeddings using t-SNE \cite{van2008visualizing} to gain insights into the differences in the embeddings between the base model and the fine-tuned model, and further explore the relationship between slang and non-slang words.

\section{Experiments and Results}

As outlined in the methodology, we calculated the cosine similarity between the embeddings of the pre-trained model and the community fine-tuned model. For each community, we identified the top 15 slang words (those with the lowest cosine similarity) and the top 15 non-slang words (those with the highest cosine similarity). Figures 1 to 3 display the slang and non-slang words for each respective community. Through this qualitative evaluation, it becomes evident that the detected slang words are unique to their respective communities. In contrast, the non-slang words are general English terms that do not possess any additional or specialized meanings within these specific communities.

Table 2 shows the training metrics of the models that we observed while fine-tuning the baseline on different community data. The table includes the metrics for both epoch 1 and epoch 2 in the fine-tuning process. The perplexity of the fine-tuned models has been observed to be significantly lower than the perplexity of the baseline community on the validation data. The perplexity of the baseline model on the validation data of the WoW subreddit is 1182.95, whereas this drops to 18.76 on fine-tuning. This supports our hypothesis that fine-tuned model is better at understanding the community text in all the communities.

Table 1 contains the F1 metrics on the fine-tuned and baseline models on the evaluation dataset. This table shows that neither the baseline nor our model performs well if we consider the F1 score as the evaluation metric for this task. However, F1 score may not even be the best metric for this task as it penalizes all wrong answers similarly despite the answers being close to what is expected. Thus how "surprised" our model is, i.e. perplexity is a better metric for this task, which our model has shown to be significantly better at than the baseline.

The table in Figure 4 showcases some essential qualitative findings from the experiment. The primary objective is to examine specific test sentences from diverse communities, perform the masked language modeling task on each sentence using different models, and subsequently compare the outcomes. As illustrated in the table, although the predictions generated by the fine-tuned models do not precisely match the ground truth in most instances, the predicted words demonstrate a closer semantic relationship to the gold labels compared to those of the base model. It is also worth noting that the "gaming" model exhibits a satisfactory performance on the "world of warcraft" sentences, signifying a degree of overlap between the two communities. This observation highlights the potential for transfer learning between related communities in our methodology.

\begin{table}[]
\begin{tabular}{llr}
data & model & f1 score \\
\hline
WOW                          & WOW DistilRoBERTa                & 0.019                                      \\
WOW                          & DistilRoBERTa                          & 0.064                    \\
Gaming                       & Gaming DistilRoBERTa             & 0.020                   \\
Gaming                       & DistilRoBERTa                          & 0.020                                                \\
Tech                         & Tech DistilRoBERTa               & 0                                           \\
Tech                         & DistilRoBERTa                          & 0                                         
\end{tabular}
\caption{\label{f1-table}F1 score for the fine-tuned and baseline models on the evaluation data}
\end{table}

\begin{table*}
\begin{tabular}{lrrrrrrr}
Model Name               & \multicolumn{1}{l}{Epochs} & \multicolumn{1}{l}{Batch} & \multicolumn{1}{l}{lr} & \multicolumn{1}{l}{weight\_decay} & \multicolumn{1}{l}{perplexity} & \multicolumn{1}{l}{training loss} & \multicolumn{1}{l}{validation loss} \\
\hline
WOW DistilRoBERTa        & 1                          & 32                        & 1.00E-04               & 0.001                             & 18.76                          & 3.335                             & 2.945                               \\
WOW DistilRoBERTa        & 2                          & 32                        & 1.00E-04               & 0.001                             & 19.05                          & 2.9083                            & 2.919814                            \\
Technology DistilRoBERTa & 1                          & 32                        & 1.00E-04               & 0.001                             & 20.02                          & 3.4011                            & 3.001113                            \\
Technology DistilRoBERTa & 2                          & 32                        & 1.00E-04               & 0.001                             & 19.92                          & 2.9436                            & 3.008951                            \\
Gaming DistilRoBERTa     & 1                          & 32                        & 1.00E-04               & 0.001                             & 12.2                           & 2.6231                            & 2.764                               \\
Gaming DistilRoBERTa     & 2                          & 32                        & 1.00E-04               & 0.001                             & 12.18                          & 2.4758                            & 2.5037 \\
Baseline on WOW data    & -                          & -                        & -               & -                             & 1182.95                          & -                           & \\
Baseline on Technology data    & -                          & -                        & -               & -                             & 753.09                          & -                           & -\\
Baseline on Gaming data    & -                          & -                        & -               & -                             & 747.70                          & -                           & -

\end{tabular}
\caption{\label{metrics-table}Training and validation metrics for fine-tuned models.}
\end{table*}

\begin{figure}[h]
  \centering
  \includegraphics[width=\linewidth]
  {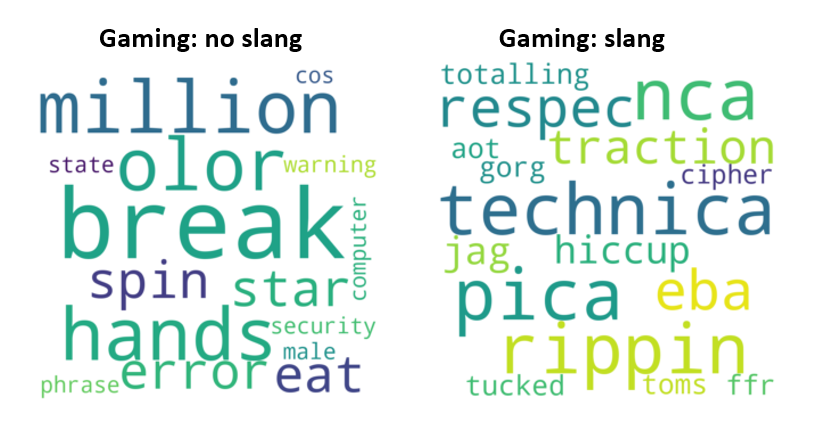}
  \caption{Gaming community}
\end{figure}

\begin{figure}[h]
  \centering
  \includegraphics[width=\linewidth]{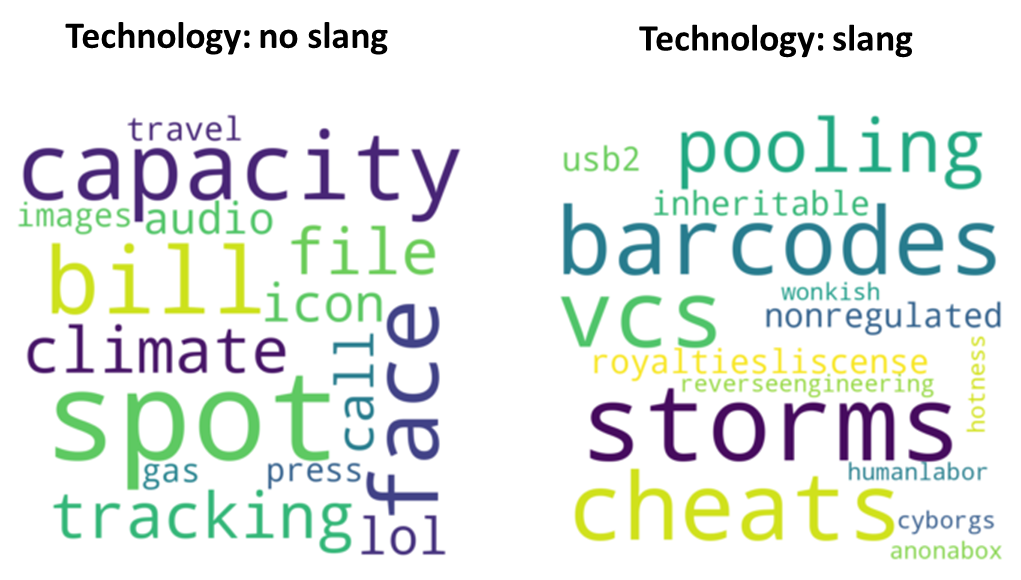}
  \caption{Technology community}
\end{figure}

\begin{figure}[h]
  \centering
  \includegraphics[width=\linewidth]{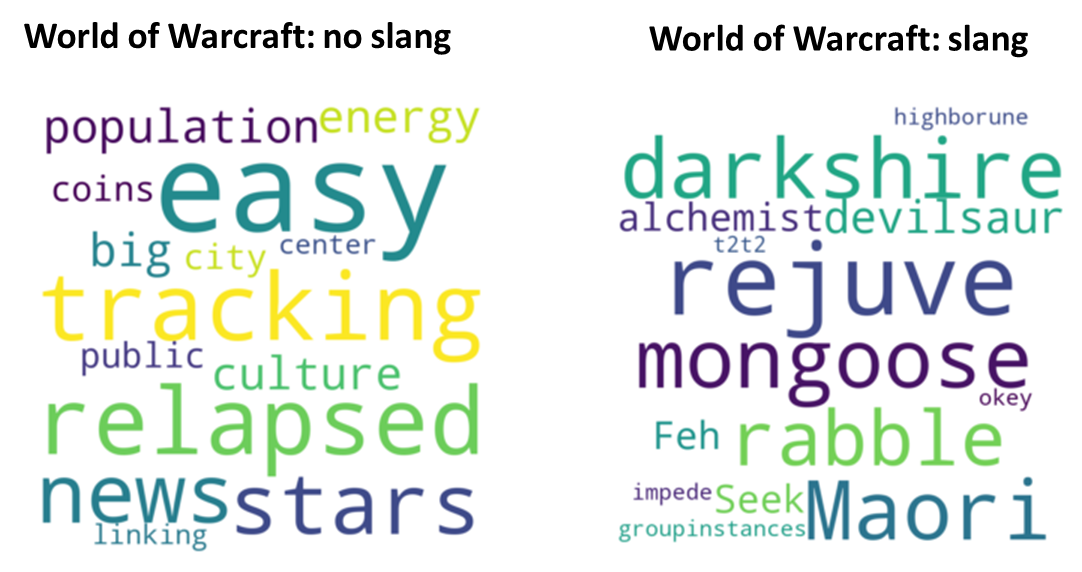}
  \caption{World of Warcraft}
\end{figure}

\begin{figure*}[h]
  \centering
  \includegraphics[width=\textwidth]{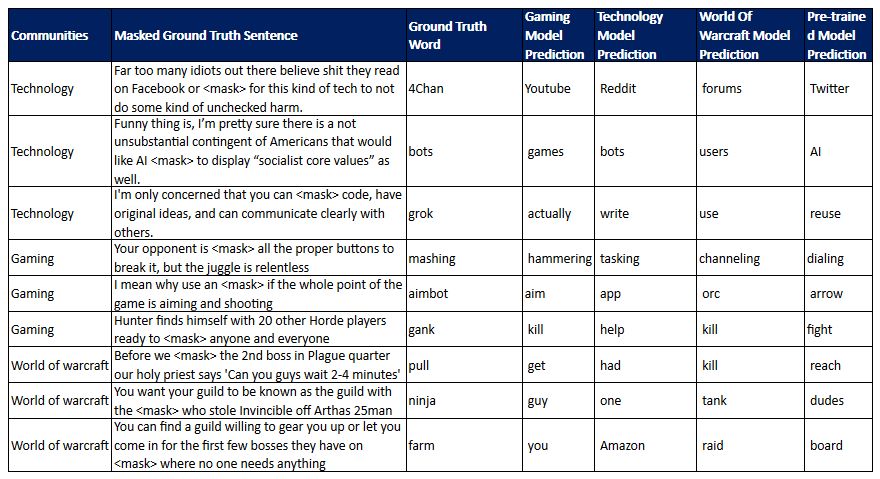}
  \caption{Qualitative comparison of models}
\end{figure*}

\begin{figure}[h]
\includegraphics[width=\linewidth]{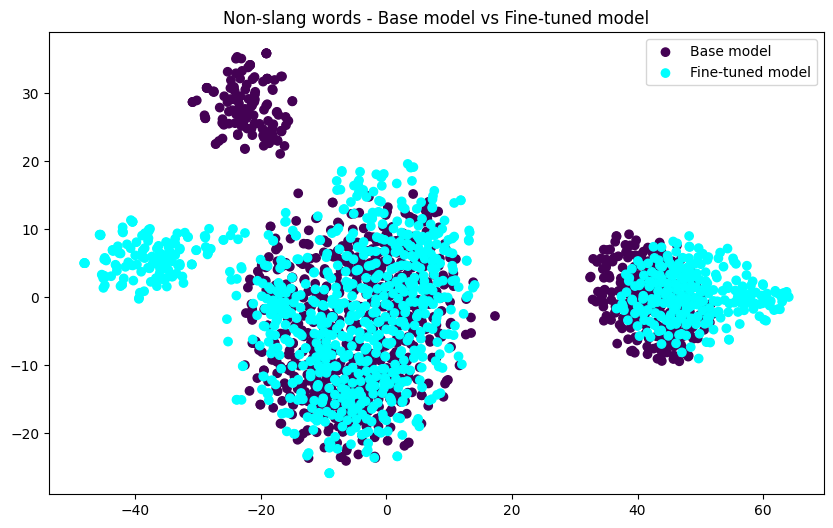}
  \caption{t-SNE visualization for non-slang word embeddings in baseline and fine-tuned model}
\end{figure}


\section{Conclusion}
Our study aimed to test the hypothesis that fine-tuning a model on a community-specific dataset would enable the identification of slang, unique entities, and folklore from that community by examining words exhibiting the most significant semantic shifts in their embedding representations. Our experiments successfully supported this hypothesis. Furthermore, we observed that slang shared between similar communities (e.g., Gaming and World of Warcraft) could be detected and translated across domains, suggesting a degree of transferability in slang detection across related communities. We found that our method drastically improved the perplexity on the validation and evaluation dataset.

\subsection{Limitations}
Due to computational limitations, we were only able to explore relatively smaller communities. Communities with larger amounts of data, like (League of Legends) may have provided further validation to our methodology, improved performance, and could be explored in future work.

We were also able to fine-tune for only 1-2 epochs, as we did not have access to higher computational resources. It was also difficult to quantitatively evaluate the fine-tuned model's ability to detect specific slang, entities, and lore within the community, as a standard benchmark for this task doesn't exist. Thus, we created an ad-hoc benchmark for evaluation, which is much smaller due to the manual collection methodology.  

\subsection{Future Work}
In the future, we hope to take this study further and expand on the study and test the following hypothesis:

1. Adding more context to our Masked LM fine-tuning process will help our model make better predictions on masked slang/entity/folklore tokens.

2. Investigate how words with high magnitude of semantic shift can be used to detect communities that have common users or share similar values/ideas/folklore/experiences by identifying communities that show a similar drift in language.

3. Having more time to fine-tune our model and expand our vocabulary can help us perform masked language modeling better.

4. This method can be expanded further to de-bias current hate speech detectors and create systems that are more context-aware.

\begin{acks}
We thank Professor Alan Ritter and Aman Khullar for their guidance and support.

\end{acks}

\printbibliography

@String{Springer = "Springer-Verlag" }

@article{liu2019roberta,
  title={Roberta: A robustly optimized bert pretraining approach},
  author={Liu, Yinhan and Ott, Myle and Goyal, Naman and Du, Jingfei and Joshi, Mandar and Chen, Danqi and Levy, Omer and Lewis, Mike and Zettlemoyer, Luke and Stoyanov, Veselin},
  journal={arXiv preprint arXiv:1907.11692},
  year={2019}
}

@article{liu2021statistically,
  title={Statistically significant detection of semantic shifts using contextual word embeddings},
  author={Liu, Yang and Medlar, Alan and Glowacka, Dorota},
  journal={arXiv preprint arXiv:2104.03776},
  year={2021}
}

@article{del2018semantic,
  title={Semantic variation in online communities of practice},
  author={Del Tredici, Marco and Fern{\'a}ndez, Raquel},
  journal={arXiv preprint arXiv:1806.05847},
  year={2018}
}

@inproceedings{pei2019slang,
  title={Slang detection and identification},
  author={Pei, Zhengqi and Sun, Zhewei and Xu, Yang},
  booktitle={Proceedings of the 23rd Conference on Computational Natural Language Learning (CoNLL)},
  pages={881--889},
  year={2019}
}

@article{wevers2020digital,
  title={Digital begriffsgeschichte: Tracing semantic change using word embeddings},
  author={Wevers, Melvin and Koolen, Marijn},
  journal={Historical Methods: A Journal of Quantitative and Interdisciplinary History},
  volume={53},
  number={4},
  pages={226--243},
  year={2020},
  publisher={Taylor \& Francis}
}

@article{van2008visualizing,
  title={Visualizing data using t-SNE.},
  author={Van der Maaten, Laurens and Hinton, Geoffrey},
  journal={Journal of machine learning research},
  volume={9},
  number={11},
  year={2008}
}

@inproceedings{sap-etal-2019-risk,
    title = "The Risk of Racial Bias in Hate Speech Detection",
    author = "Sap, Maarten  and
      Card, Dallas  and
      Gabriel, Saadia  and
      Choi, Yejin  and
      Smith, Noah A.",
    booktitle = "Proceedings of the 57th Annual Meeting of the Association for Computational Linguistics",
    month = jul,
    year = "2019",
    address = "Florence, Italy",
    publisher = "Association for Computational Linguistics",
    url = "https://aclanthology.org/P19-1163",
    doi = "10.18653/v1/P19-1163",
    pages = "1668--1678",
}

@article{camille-2021,
   title={Mitigating Racial Biases in Toxic Language Detection with an Equity-Based Ensemble Framework},
   url={http://dx.doi.org/10.1145/3465416.3483299},
   DOI={10.1145/3465416.3483299},
   journal={Equity and Access in Algorithms, Mechanisms, and Optimization},
   publisher={ACM},
   author={Halevy, Matan and Harris, Camille and Bruckman, Amy and Yang, Diyi and Howard, Ayanna},
   year={2021},
   month={Oct} }

@article{deepinf-2018,
   title={DeepInf},
   url={http://dx.doi.org/10.1145/3219819.3220077},
   DOI={10.1145/3219819.3220077},
   journal={Proceedings of the 24th ACM SIGKDD International Conference on Knowledge Discovery \& Data Mining},
   publisher={ACM},
   author={Qiu, Jiezhong and Tang, Jian and Ma, Hao and Dong, Yuxiao and Wang, Kuansan and Tang, Jie},
   year={2018},
   month={Jul} }

@article{bagrow-2019,
   title={Information flow reveals prediction limits in online social activity},
   volume={3},
   ISSN={2397-3374},
   url={http://dx.doi.org/10.1038/s41562-018-0510-5},
   DOI={10.1038/s41562-018-0510-5},
   number={2},
   journal={Nature Human Behaviour},
   publisher={Springer Science and Business Media LLC},
   author={Bagrow, James P. and Liu, Xipei and Mitchell, Lewis},
   year={2019},
   month={Jan},
   pages={122–128} }

\end{document}